\pdfoutput=1
\documentclass[runningheads]{llncs}
\usepackage{graphicx}
\usepackage{amsmath} % assumes amsmath package installed
\usepackage[caption=false]{subfig} %for subfigures
\usepackage{url}
\begin{document}
\title{Multimodal Contact Detection using Auditory and Force Features for Reliable Object Placing in Household Environments}

\titlerunning{Multimodal Contact Detection for Reliable Object Placing}%Abbreviated paper title
% If the paper title is too long for the running head, you can set
% an abbreviated paper title here
%
%Ca\~{n}\'{o}n
\author{Jaime L. Maldonado C.\orcidID{0000-0002-2334-1073} \and  Asil Kaan Bozcuo\u{g}lu
\and Christoph Zetzsche %\and
%Third Author\inst{3}\orcidID{2222--3333-4444-5555}
}
\authorrunning{J. Maldonado et al.}
%\authorrunning{F. Author et al.}
% First names are abbreviated in the running head.
% If there are more than two authors, 'et al.' is used.
%
\institute{Cognitive Neuroinformatics and Institute for Artificial Intelligence, University of Bremen, Bremen, Germany 
\email{jmaldonado@uni-bremen.de, asil@uni-bremen.de, zetzsche@informatik.uni-bremen.de} \\
}
\maketitle              % typeset the header of the contribution
%
%%%%%%%%%%%%%%%%%%%%%%%%%%%%%%%%%%%%%%%%%%%%%%%%%%%%%%%%%%%%%%%%%%%%%%%%%%%%%%%%
\begin{abstract}
	Typical contact detection is based on the monitoring of a threshold value in the force and torque signals. The selection of a threshold is challenging for robots operating in unstructured or highly dynamic environments, such in a household setting, due to the variability of the characteristics of the objects that might be encountered. We propose a multimodal contact detection approach using time and frequency domain features which model the distinctive characteristics of contact events in the auditory and haptic modalities. In our approach the monitoring of force and torque thresholds is not necessary as detection is based on the characteristics of force and torque signals in the frequency domain together with the impact sound generated by the manipulation task. We evaluated our approach with a typical glass placing task in a household setting. Our experimental results show that robust contact detection (99.94\% mean cross-validation accuracy) is possible independent of force/torque threshold values and suitable of being implemented for operation in highly dynamic scenarios. 

\end{abstract}

%%%%%%%%%%%%%%%%%%%%%%%%%%%%%%%%%%%%%%%%%%%%%%%%%%%%%%%%%%%%%%%%%%%%%%%%%%%%%%%%
\section{Introduction}
One of the major problems in autonomous robotics is to operate in unstructured, highly dynamic or unknown environments including the inherent challenge to react to unexpected situations during task execution \cite{Haddadin_etal_2017}. The ability to appropriately react to such unexpected events requires to continuously monitor the task progression. There is a variety of situations household robots could be confronted with, such as placing objects or using tools, and one typical single element that exists in most of these action sequences is the \emph{contact event}. 

%about contacts in different sensor modalities:
Contact events that happen during the execution of everyday activities produce characteristic sensory information in the visual, haptic and auditory modalities. Haptic information includes the changes in the load force and torques experienced at hand-held tools or manipulated objects. Regarding the acoustic information, discrete and continuous interactions, such as tapping or rubbing a surface with a tool, produce distinctive sounds which can be identified as the acoustic signature of an action-object interaction.

%the importance/motivation for a multimodal approach
In order to make use of the sensory information of the individual modalities there is a number of potential uncertainties that must be considered during the design of contact detection systems. These uncertainties are related not only to the robot's perception capabilities but also to the characteristics of the manipulated objects and the strategy followed by the robot to execute the task. The processing of multimodal sensory information in robotic applications aims to compensate the uncertainties or shortcomings of the individual sensor modalities. Monitoring multimodal signals during task execution enables the implementation of different functionalities such as task monitoring (e.g. for success-failure detection) \cite{Park_etal_2016}, behavior switching \cite{Park_etal_2016}, modeling the effects of different actions relevant for the task \cite{Chu_etal_2019} and action adaptation \cite{Chu_etal_2019}. 

Detection of contacts and collisions (i.e. unintended contacts) is typically focused on the haptic modality and is implemented by monitoring thresholds values in the force and torque signals  \cite{Haddadin_etal_2017}\cite{Haninger_Surdilovic_2018} or by monitoring filtered force/torque signals \cite{Cho_etal_2012} \cite{Li_etal_2019}. In this paper we present a multimodal contact detection approach which, instead of monitoring signals in the time domain, uses time and frequency domain features which model the distinctive characteristics of contact events in the auditory and haptic modalities. In particular, our approach addresses the uncertainties inherent to the force and torque sensor readings that depend on the characteristics of the manipulated object (such as weight and dimensions) and the manipulation task. In the context of a household setting these uncertainties pose an enormous difficulty to achieve a robust contact detection during the manipulation of diverse objects, including unknown objects and objects with dynamic characteristics (e.g. liquid containers where the amount of liquid is unknown). The advantages of our approach can be summarized as follows: \textit{(i)} we do not rely on the monitoring of force/torque threshold values which are difficult to determine a priori, \textit{(ii)} we do not rely on position information \textit{(iii)} we do not rely on velocity or acceleration information which can be difficult to determine in real time. 

\begin{figure}[thpb]
	\centering
	\framebox{\parbox{3.2in}{
			\includegraphics[scale=0.057]{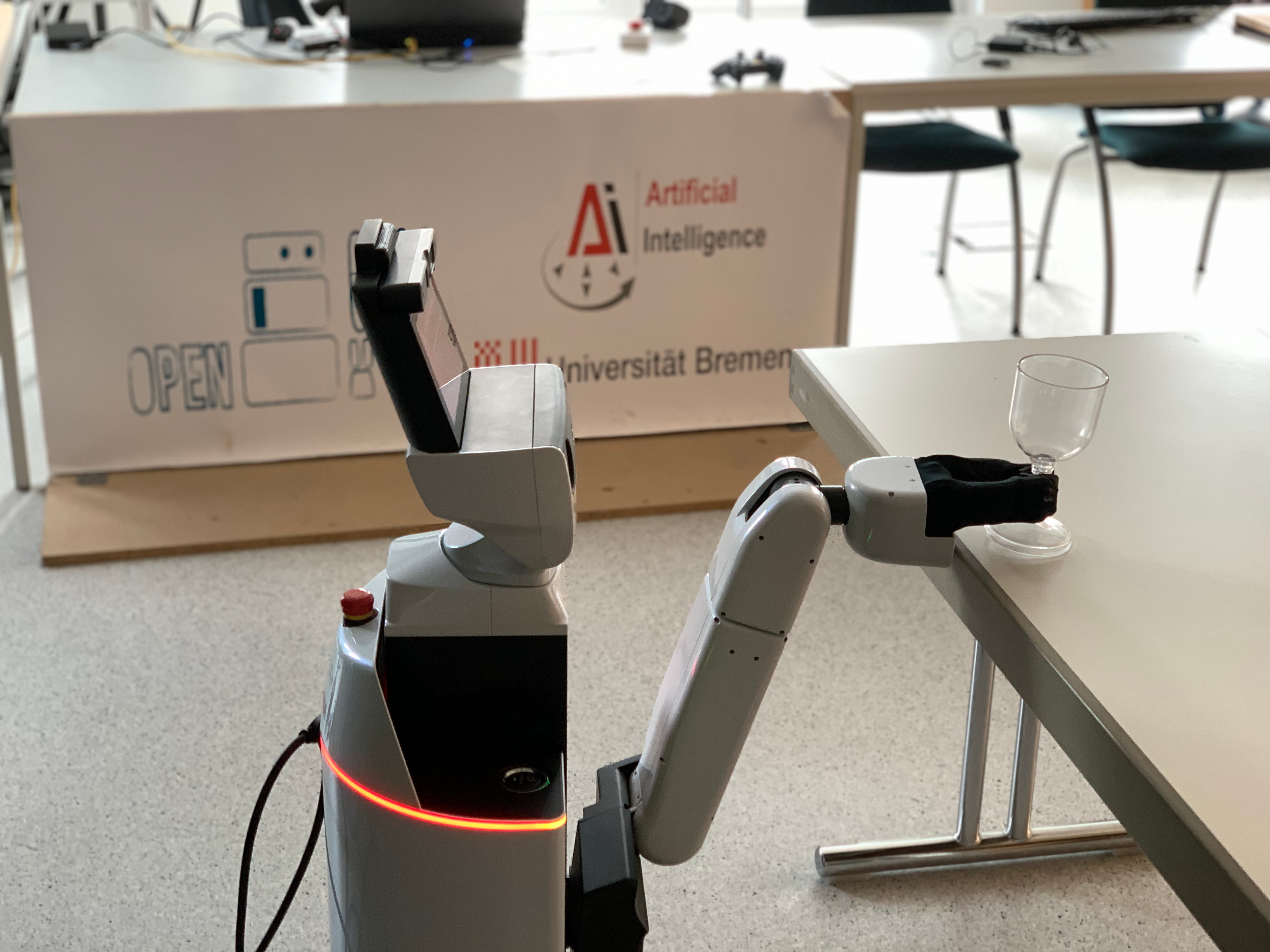}
	}}
	
	\caption{Experimental setup: robot placing the glass on the table. The robot is equipped with a built-in head microphone and a force/torque sensor in the wrist.}
	\label{fig:HSR_robot}
\end{figure}

We evaluated our contact detection approach with a prototypical glass placing task in a household setting. The experimental setup is shown in Figure \ref{fig:HSR_robot}. Our experimental results show that robust contact detection is possible independent of the monitoring of force/torque threshold values and indicate that the selected features provide a good representation of the sensory signature of contact events, and thus are well suited for contact detection applications.

\section{Related Work} \label{sec:related_work}

In this section we focus on recent examples of multimodal task execution and monitoring applications together with the properties of sound and force signals relevant for the detection of contacts. 

\subsection{Audio information in multimodal task execution and monitoring applications}

% % % % % % % % % % % % % % % % % % % % % % % % % % % % % % % % % % % %
%ANOMALY - FAULT DETECTION

Park \textit{et al}. \cite{Park_etal_2016} used the relation between force and sound for execution monitoring and anomaly detection. They detected anomalies using haptic and auditory signals during the execution of a pushing task in which the robot closed a microwave oven. In their approach, anomaly detection is based on the comparison of the current signals with signal features associated with successful task executions, which are characterized by a stereotypical force and sound energy profile (loud sound and force reduction when the door is secured). 
    
% % % % % % % % % % % % % % % % % % % % % % % % % % % % % % % % % % %
%(MULTI MODAL) ACTION ADAPTATION

Chu \textit{et al}. \cite{Chu_etal_2019} investigated action adaptation to manipulate novel objects (i.e. not previously encountered) and adaptation to changes in previously learned objects. Their approach is based on the adaptation of an \textit{affordance model} of different object-action pairs in the haptic, audio, and visual modalities. They collected interactions of a robot opening a drawer and turning on a lamp. With these data they modeled action-object pairs during for different segments of task execution (e.g. grasp and release segments). The adaptation to changes in previously learned objects was evaluated by letting a robot to open a drawer in different states, ranging from fully closed to fully open. Manipulation of novel objects was evaluated by letting the robot to turn on different lamps, which differed in shape, size and length of pull chain, in which the robot had to determine the point to stop pulling the chain. 

\subsection{Properties of impact sounds} \label{subsec:properties_impact_sounds}
Impact sounds occur during the manipulation of objects and tools in everyday life. The characteristics of the sound depend on the objects interacting (e.g. when placing a glass on a wood table) and the movement that generated the sound (e.g. a soft or harsh move). Thus impact sounds encode perceptual information related to the physical attributes of the objects interacting (material, shape, size) and the movement (impact force) \cite{Cook_97}\cite{vandenDoel_Pai_2007}. Impact sounds are characterized by a short duration, an abrupt onset and a rapid decay \cite{Aramaki_Kronland_2006}. These particular characteristics are taken into account in the field of audio analysis and computer audition. In particular, features based on the energy level of the audio signal can be used for the detection and classification of sounds characterized by significant and abrupt energy changes such as gunshots, explosions or other environmental sounds \cite{Giannakopoulos_Pikrakis_2014}.   
sounds, etc."  

\subsection{Detection of collisions and contacts} \label{subsec:collision_event_pipeline}
The relevant phases involved in the detection and handling of collisions in robotic systems have been reviewed by Haddadin \textit{et al.} \cite{Haddadin_etal_2017}. Common approaches for collision detection include the monitoring of the measured currents in the electrical drives and the monitoring of the instantaneous torque \cite{Haddadin_etal_2017}. A collision detection system should be fast with a minimum occurrence of false detections \cite{Haddadin_etal_2017}. The major design challenge is the selection of a threshold on the monitoring signals to avoid false positives and to achieve high sensitivity. The selection of a monitoring threshold in any of the common approaches is difficult because of the highly varying dynamic characteristics of the control torques \cite{Haddadin_etal_2017}. Furthermore, a robust detection system should take the following factors into account which depend on the robot state, load torque, temperature, and time \cite{Haddadin_etal_2017}: torque/current measurement noise, position and velocity sensor noise and modeling errors in the estimated robot dynamics. Haddadin \textit{et al.}  \cite{Haddadin_etal_2017} pointed out the similarities of collisions and impacts during robot manipulation tasks, thus the relevant aspects of collision detection also apply to object and tool manipulation scenarios.

%%%%%%%%%%%%%%%%%%%%%%%%%%%%%%%%%%%%%%%%%%%%%%%%
%FORCE SIGNAL IN THE FREQUENCY DOMAIN
\subsection{Properties of force/torque signals during contacts and collisions in the frequency domain} \label{subsec:force_in_freq_domain}
In the field human-robot interaction the characteristics of the force and torque signals in the frequency domain have been considered for the detection of contacts and collisions. Cho \textit{et al.}  \cite{Cho_etal_2012} implemented a collision detection algorithm based on the observation that the force during an unintended collision has a faster rate of change compared to the rate measured during intended contacts. The torque data is monitored by means of a high-pass filter which enables the distinction between intended contacts and unexpected collisions. Following a similar approach, Li \textit{et al.} \cite{Li_etal_2019} implemented a low-pass and band-pass filter observer for robot contact and collision detection. Since the frequency components of the  collision  force  signals are located in higher bands than those of the intentional contact force signals, the two filters enable the distinction between contacts and collisions.

\section{Problem definition}
%the task and how it is performed
The task we are dealing with is placing objects on a surface in the context household everyday activities. The properties of the manipulated objects differ due to various shapes, sizes and weights available. In this context, the robot might be requested, for example, to set dishes and glasses on the table for dinner.

The robot estimates the distance between the in-hand object and the support surface and then is commanded beyond a worst-case position. A force threshold is usually used to detect the contact of the object against the surface. The object can be released once contact has been detected. 

\begin{figure}[thpb]
	\centering
	\framebox{\parbox{3.2in}{
		\includegraphics[scale=0.650]{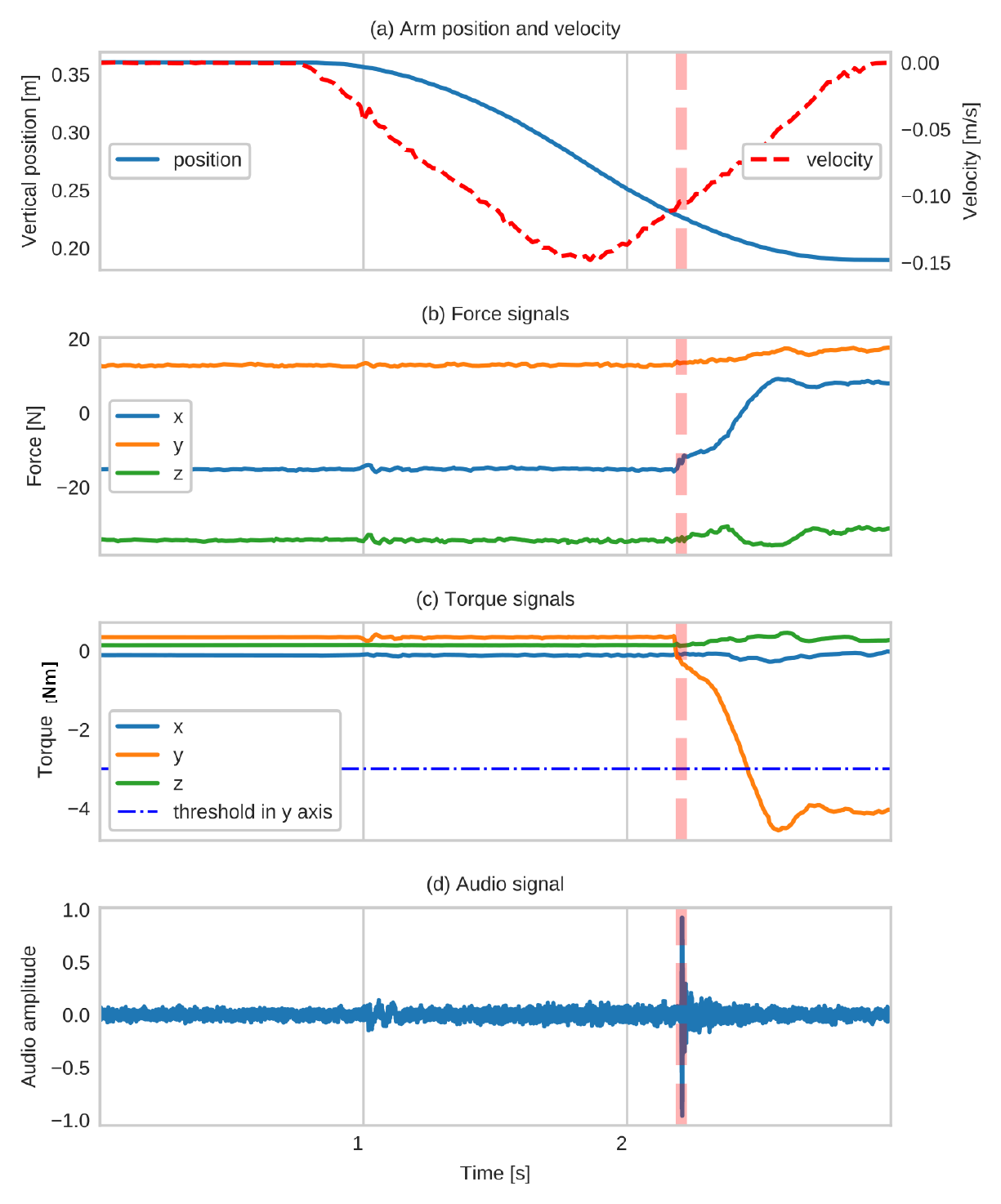}
		}}
		
		\caption{Multimodal signals as the robot places the glass on the table. \textbf{(a)} Position and velocity. \textbf{(b)} 3-axis force signals. \textbf{(c)} 3-axis torque signals and torque threshold used to detect the contact of the glass on the table. \textbf{(d)} Audio signal. Apart from the impact sounds, the audio signal also displays the ego noise of the robot and the background noise of the laboratory. The dashed semi-transparent vertical line indicates the moment in which the bottom of the glass makes contact with the table.}
		\label{fig:1_signals}
	\end{figure}

Figure \ref{fig:1_signals} illustrates the time course of the position, force, torque and audio signals registered during the manipulation task. The initial force magnitude corresponds to the weight of the grasped object. The initial amplitude of the audio signal corresponds to the background noise of the environment and the robot's ego noise. At the beginning of the movement a small perturbation of the baseline measured force can be observed. Correspondingly, the sound of the robot's actuators can be observed. At the moment of contact the characteristic abrupt onset of the audio signal and its rapid decay can be observed together with the change in the load force and the torque registered at the robot's sensor. After the glass makes contact with the surface the robot arm keeps on moving toward the commanded position until the torque threshold is detected, indicating that the goal of the task has been accomplished.

Detection of contacts is typically realized by monitoring hard-coded thresholds in the  force signal \cite{Haddadin_etal_2017}\cite{Haninger_Surdilovic_2018}. While this strategy can work in well structured environments where the characteristics of the manipulated objects are well known, the establishment of suitable thresholds in unstructured environments is challenging due to the sources of uncertainty of the task. Apart from the uncertainties related to the robot and its sensors listed in section \ref{subsec:collision_event_pipeline}, the weight, shape and dimensions of the manipulated object influence the force/torque measured values. In particular, object weights can be difficult to determine \textit{a priori}, as in the case of liquid containers. Furthermore, certain objects might be subject to contact force constraints which can be encountered during the manipulation of fragile objects. In this case, the unnecessary exertion of force after contact might risk the object's integrity.

\section{Methodology}
\begin{figure}[thpb]
	\centering
	\framebox{\parbox{3.2in}{
			\includegraphics[scale=0.55]{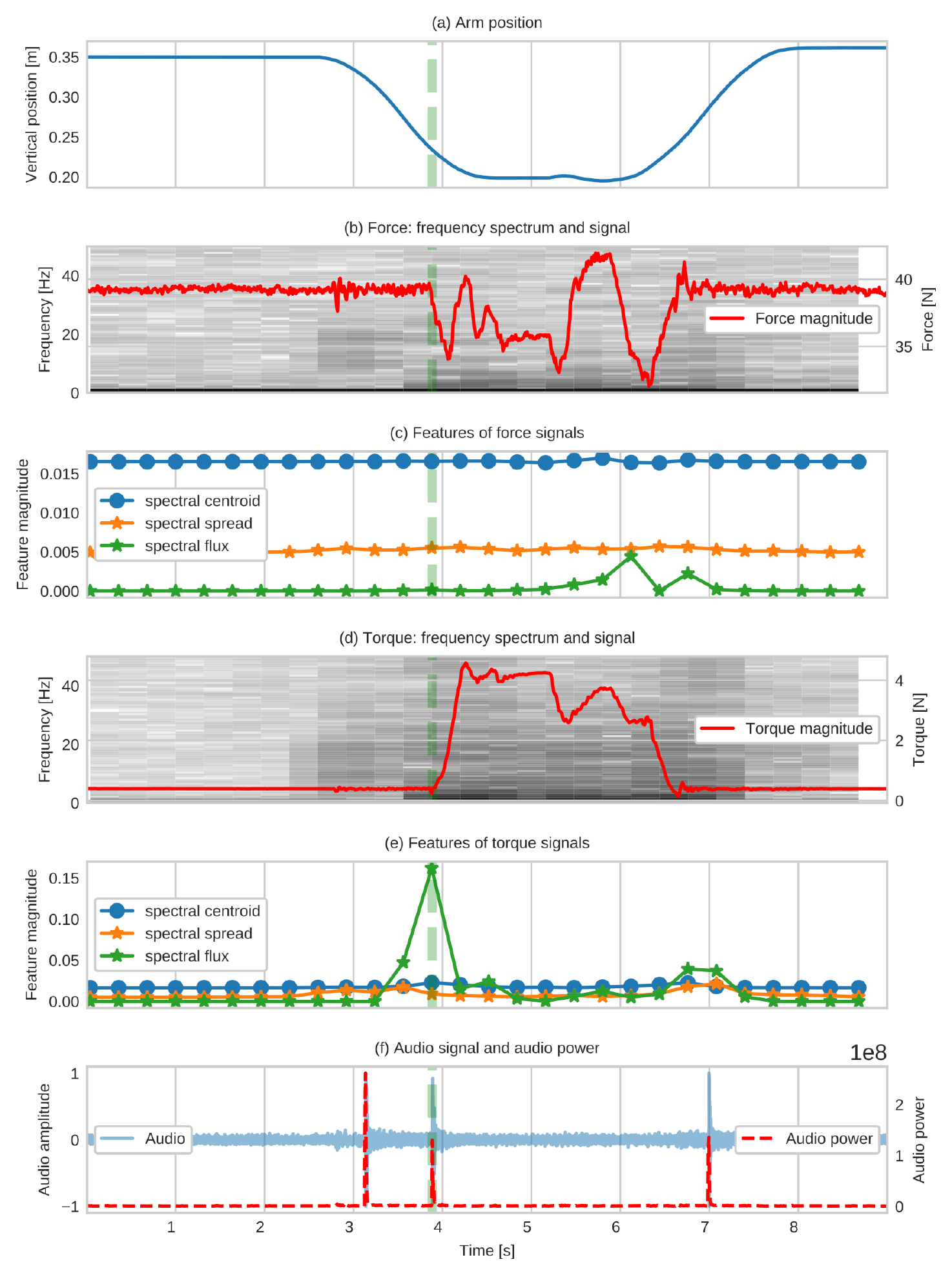}
	}}
	
	\caption{Multimodal signals and features during the manipulation task. \textbf{(a)} Position. The robot is commanded toward the position of the table. After the torque threshold is detected the robots moves its arm toward the start position. \textbf{(b)} Magnitude of the force signal and its spectrogram. \textbf{(c)} Frequency domain features computed from the force signal. \textbf{(d)} Magnitude of the torque signal and its spectrogram. \textbf{(e)} Frequency domain features computed from the torque signal. \textbf{(f)} Audio signal and its corresponding audio power feature. The dashed semi-transparent vertical line indicates the moment in which the bottom of the glass makes contact with the table. Subfigures (b) and (d) illustrate the change of the in the frequency content of the force and torque signal at the initial contact and during the sustained contact of the glass with the table. Subfigure (f) illustrates two exogenous contact sound that ocurred during task execution (visible as peaks in the curve of the audio feature). }
	\label{fig:example_features}
\end{figure}

Figure \ref{fig:example_features} illustrates the force and sound signals during the manipulation task. We use these signals to obtain time and frequency domain features for contact detection. As described in section \ref{subsec:force_in_freq_domain}, force and torque signals show a specific behavior in the frequency domain during contact. This behavior is illustrated in the spectrograms shown in Figure \ref{fig:example_features}. Before the movement onset, the signal shows a strong low frequency component. It can also be observed the amplitude of the frequency components between 0 Hz and 25 Hz  increase at movement onset and at the moment of contact. This behavior can be easily observed in the torque signal. As opposed to the filter-based detection approaches used in \cite{Cho_etal_2012} and \cite{Li_etal_2019}, we propose to measure the changes in the frequency domain of the force and torque signals by means of spectral features, which will be described in section \ref{subsec:force_features}. 

The sound signal produced at the moment of contact is illustrated in Figure \ref{fig:example_features}. As described in section \ref{subsec:properties_impact_sounds}, contact sounds can be modeled by means of features based on the energy level of the audio signal. The time-domain feature used to identify contact sounds is described in section \ref{subsec:audio_feature}.   

The relation between the force, torque, audio signals and their corresponding features can be observed in Figure \ref{fig:example_features}. In order to avoid the shortcomings and the challenges of using detection procedures based on position, velocity, and acceleration data or force threshold values, we propose to detect contacts based on frequency domain force features and time domain audio features. In order to detect contacts from the multimodal features we trained a \textit{Random Forest} classifier. The characteristics of the classifier and its parameters are described in section \ref{subsec:classifier}. 
 
\subsection{Experimental setup and task description}
%robot description
For the evaluation of our multimodal contact detection approach we used a Toyota Human Support Robot (HSR) (Fig. \ref{fig:HSR_robot}). The robot is designed to support human household activities and to provide assistance for handicapped persons. The HSR has a single arm with 5 DoF attached to a gripper equipped with tactile sensors and suction capability. The wrist is equipped with a 3-axes force/torque sensor sampled at 100 Hz. The HSR's base has omni-directional movement capabilities and can be lifted by a prismatic joint (movable range 0-1,350mm). For sound sensing, the HSR employs a Playstation 3 Eye Microphone mounted at the top of its head. The raw audio signal from the HSR's microphone was recorded at 44.1 kHz.

We recorded 60 task executions of the robot placing a glass on a table as shown in Figure \ref{fig:HSR_robot}. After grasping the glass, the robot was commanded to lift its arm to a start position of 36 cm. In order to put the glass on the table, the robot was commanded to move downwards and to monitor the torque signal in the y axis. The torque threshold was set at -3 N. After the contact of the glass with the table the robot kept moving downwards until the magnitude of torque in the y axis exceeded the threshold. At this point the robot was commanded to move its arm upwards. The trial ended once the robot returned to the start position. The course of the trial enabled the recording of acoustic and force/torque signals at rest (at the beginning and end of each trial), during movement onset and offset, at the moment of the initial contact and during the sustained contact between the glass and the table.
 
In order to imitate a real-life scenario in which impact sounds from other sources might be registered in the auditory modality, we added two types of exogenous impact sounds during the movement execution. In 20 trials the experimenter was hitting randomly the robot's hand and in 20 trials the experimenter was hitting randomly the table.

\subsection{Time domain audio feature}\label{subsec:audio_feature}
%AUDIO FEATURES - AUDIO PROCESSING
Features based on the energy level of the audio signal can be used for the detection and classification of sounds characterized by significant and abrupt energy changes \cite{Giannakopoulos_Pikrakis_2014}, such as those produced during the initial contact of the glass and the table.    

In order to monitor the abrupt changes in the audio signal, the normalized power $P(i)$ of the signal was computed: 

$$
P(i) = \dfrac{1}{W_{L}} \sum_{n=1}^{W_{L}} |x_{i}(n)|^{2} \eqno{(1)}
$$

in which $x_{i}(n)$ is the sequence of audio samples of the \textit{ith} frame with $ n=1 , \dots, W_{L} $ , where $W_{L}$ is the length of the frame. The power is normalized by dividing it with $W_{L}$ in order to remove the dependency on the frame length \cite{Giannakopoulos_Pikrakis_2014}. The frame length was 512 samples and the hop window 160 samples. The audio signal processing was implemented in HARK, an open-source robot audition software \cite{Nakadai_etal_2010}, and $P(i)$ was computed with a custom python script executed in a HARK processing node.

\subsection{Frequency domain features of the force and torque signals}\label{subsec:force_features}
As shown in section \ref{subsec:force_in_freq_domain}, force and torque signals show a specific behavior in the frequency domain during contacts. In order to measure the changes of the force and torque signals during the task execution we computed the \textit{spectral centroid}, the \textit{spectral spread} and the \textit{spectral flux} features. These features are computed on the Discrete Fourier Transform (DFT) coefficients obtained for each signal frame. In the following equations $X_{i}(k)$ is the magnitude of the DFT coefficients of the \textit{ith} signal frame with $  k=1 , \dots, W_{fL} $, where $W_{fL}$ is the number of DFT coefficients. The DFT was computed over 160 coefficients with a hop window of 128 samples.

%\subsubsection{Spectral centroid}
The spectral centroid is the 'center of gravity' of the spectrum. The spectral centroid $C_{i}$ of the \textit{ith} signal frame is defined as:

$$
C_{i} = \dfrac{ \sum_{k=1}^{W_{fL}} kX_{i}(k) }{ \sum_{k=1}^{W_{fL}} X_{i}(k) } \eqno{(2)}
$$

%\subsubsection{Spectral spread}
The spectral spread $S_{i}$, which provides a measure of how the spectrum is distributed around $C_{i}$, is defined as:   

$$
S_{i} = \sqrt{ \frac{ \sum_{k=1}^{W_{fL}} (k-C_{i})^{2} X_{i}(k) }{ \sum_{k=1}^{W_{fL}} X_{i}(k) } } \eqno{(3)}
$$	
	
$C_{i}$ corresponds to the frequency content of the signal (e.g. high $C_{i}$ values correspond to a signal with dominant components in the high frequency bands) and $S_{i}$ is commonly associated with the bandwidth of the signal (i.e. low values correspond to a spectrum tightly concentrated around the spectral centroid). Both $C_{i}$ and $S_{i}$ were normalized to the range [0, 1] by dividing their values by $F_s/2$ (where $F_s$ is the sampling rate of the force and torque signals). 

The spectral flux $Fl$ is a measure of the change of the spectrum between two successive signal frames. $Fl$ of the \textit{ith} frame is defined as:

$$
Fl_{(i, i-1)} = \sum_{k=1}^{W_{fL}} ( EN_{i}(k) - EN_{i-1}(k) )^{2} \eqno{(4)}
$$

where $EN_{i}(k) = \frac{ X_{i}(k) }{ \sum_{l=1}^{W_{fL}} X_{i}(l)} $ is the \textit{kth} normalized DFT coefficient at the \textit{ith} frame.

\subsection{Contact detection using a random forest classifier}\label{subsec:classifier}

\textit{Random forest} is a supervised machine learning algorithm which uses decision trees as its main building block \cite{Mueller_Guido_2017}. This algorithm does not require scaling of the data. The parameters of the algorithm are $n_{estimators}$ (i.e. the number of trees) and $max_{features}$, which controls the number of features that are selected in each node of the decision tree \cite{Mueller_Guido_2017}. The $max_{features}$ parameter was set to the recommended value for classification tasks $max_{features}=\sqrt{n_{features}}$  \cite{Mueller_Guido_2017}. Regarding the selection of the number of trees to build the model, a larger number of trees provides smoother and more robust decision boundaries. However, a large number of trees increases the computational cost with marginal improvements in the classification accuracy. The parameter $n_{estimators}$ was determined by comparing the classification performance with different settings (see section \ref{subsec:determination_of_n_trees}). We used the implementation of the random forest classifier available in the python \textit{sklearn} library version 0.20.3.

In order to detect the contact of the glass on the table we trained a random forests model with time and frequency domain features. The length of the feature sequences obtained from the audio and force/torque signals were different due to the differing sampling rates. Being the power of the audio signal the feature with the highest resolution, the force/torque features were interpolated to match the length of the audio feature sequence. The interpolated feature samples were labeled as \textit{contact} or \textit{not contact} and were divided into training (56\% of the data), validation (19\% of the data) and test sets (25\% of the data). The performance of the trained model with selected $n_{estimators}$ is presented in section \ref{subsec:classification_results}.

\section{Evaluation of Multimodal Contact Detection}
\begin{figure}[thpb]
	\centering
	\framebox{\parbox{3.2in}{
			\includegraphics[scale=0.55]{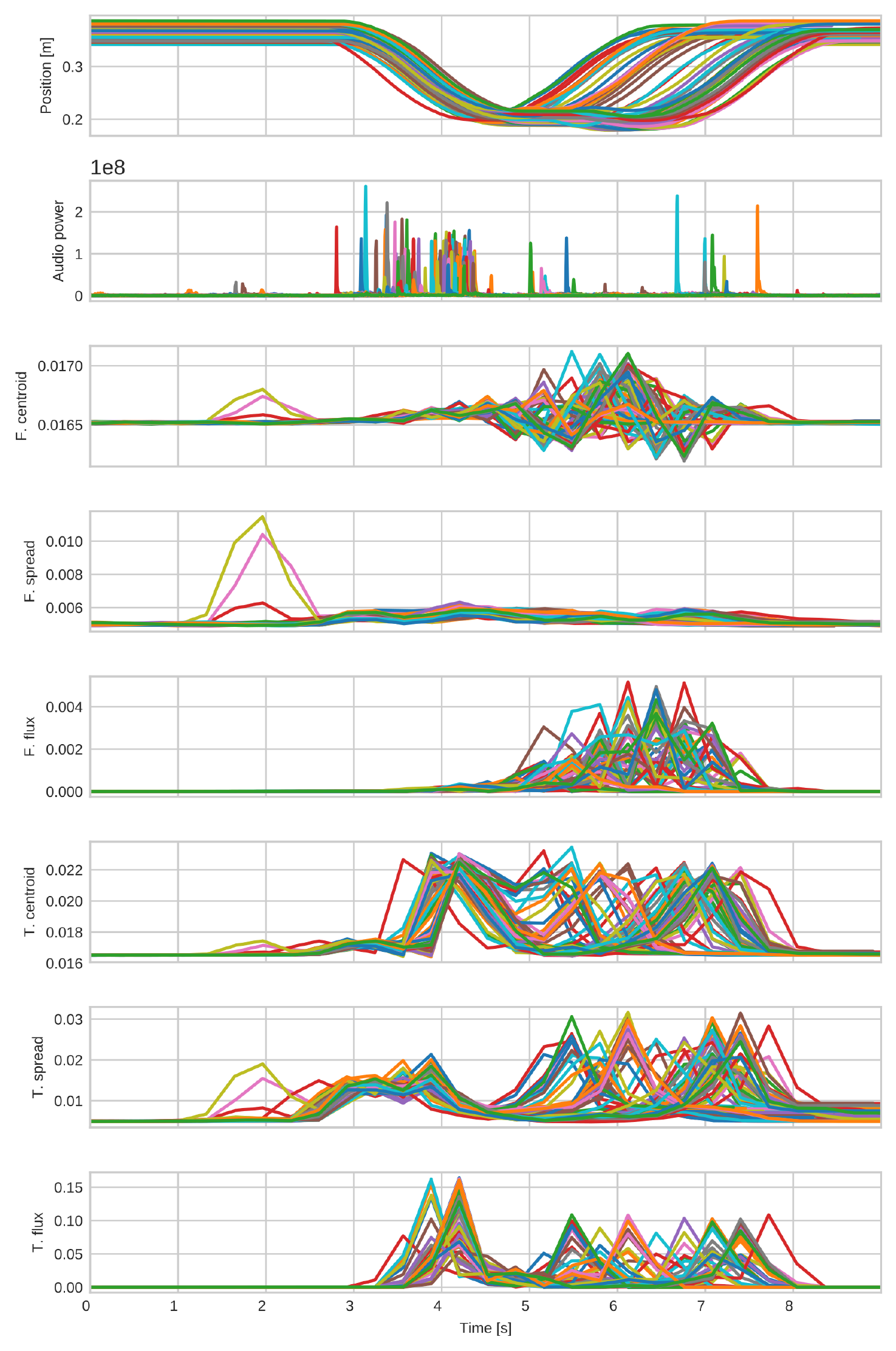}
	}}
	
	\caption{Position, audio and force (F) and torque (T) features for the 60 executions of the task.}
	\label{fig:features_over_all_trials}
\end{figure}

Figure \ref{fig:features_over_all_trials} shows the audio and force features computed for the 60 executions of the task. The peaks in the audio power subplot show the occurrence of contact and not-contact related impact sounds. By visual inspection it can be observed that force and torque features provide a good representation of the behavior of the signal during the movement onset and at the moment of contact of the glass and the table. However, it is important to notice that values of the force and torque/features at the moment of contact are similar to those measured when the robot moves its arm upwards returning to the start position. In this case, the audio power feature carries the information necessary to distinguish contacts from non contacts whenever the force/torque features show ambiguous values.   

%visualization of acquired data

\subsection{Determination of the number of trees of the random forest algorithm} \label{subsec:determination_of_n_trees}
The number of trees ($n_{estimators}$) was determined by comparing the \textit{Area Under the Curve} (AUC) score of the \textit{Receiver Operating Characteristic} (ROC) obtained with the validation set for models trained with different $n_{estimators}$ ranging from 1 to 10. The results are displayed in Table \ref{table:n_trees_validation}. The accuracy of all the $n_{estimators}$ settings is above 98\%. The AUC values, which express the average precision with a value between 0 (worst) and 1 (best) \cite{Mueller_Guido_2017}, are above 0.99 for all the settings. As recommended in \cite{Mueller_Guido_2017} for classification problems with imbalanced classes (in our dataset \textit{not contact} samples occur more often), the AUC is used as criterion for the selection of the  $n_{estimators}$ parameter. Thus the number of trees is set to $n_{estimators}=8$. 

%%%%%%%%%%%%%%%%%%%%%%%%%%%%%%%%%%%%%%%%
%VALIDATION OF THE NUMBER OF TREES

\begin{table}[h]
	\caption{Comparison of different AUC and accuracy scores with different  $n_{estimators}$}
	\label{table:n_trees_validation}
	\begin{center}
		\begin{tabular}{|c|c|c|}
			\hline
			\textbf{$n_{estimators}$} &\textbf{ accuracy} & \textbf{AUC}\\
			\hline
			1 & 0.97719 & 0.99977\\
			2 & 0.98174 & 0.99957\\
			3 & 0.99084 & 0.99972\\
			4 & 0.99084 & 0.99965\\
			5 & 0.99084 & 0.99973\\
			6 & 0.99994 & 0.99977\\
			7 & 0.99994 & 0.99978\\
			\textbf{8} &\textbf{ 0.99994} & \textbf{0.99981}\\
			9 & 0.99994 &  0.99978\\
			10 & 0.99994 &  0.99978\\
			\hline
		\end{tabular}
	\end{center}
\end{table}

\subsection{Validation of the random forest classifier}

In order to evaluate the extent to which the model trained with $n_{estimators}=8$ generalizes over the whole data set we performed an stratified 10-fold cross-validation. The mean cross-validation accuracy indicates that the trained model is 99.94\% ($SD = 0.04 \%$) accurate on average. The standard deviation indicates that there is only little variance in the accuracy between folds, thus the model is independent from the particular folds used for training.

\subsection{Classification results} \label{subsec:classification_results}

%%%%%%%%%%%%%%%%%%%%%

\begin{figure}[thpb]
	\centering
	\framebox{\parbox{2,2in}{
			\includegraphics[scale=0.5]{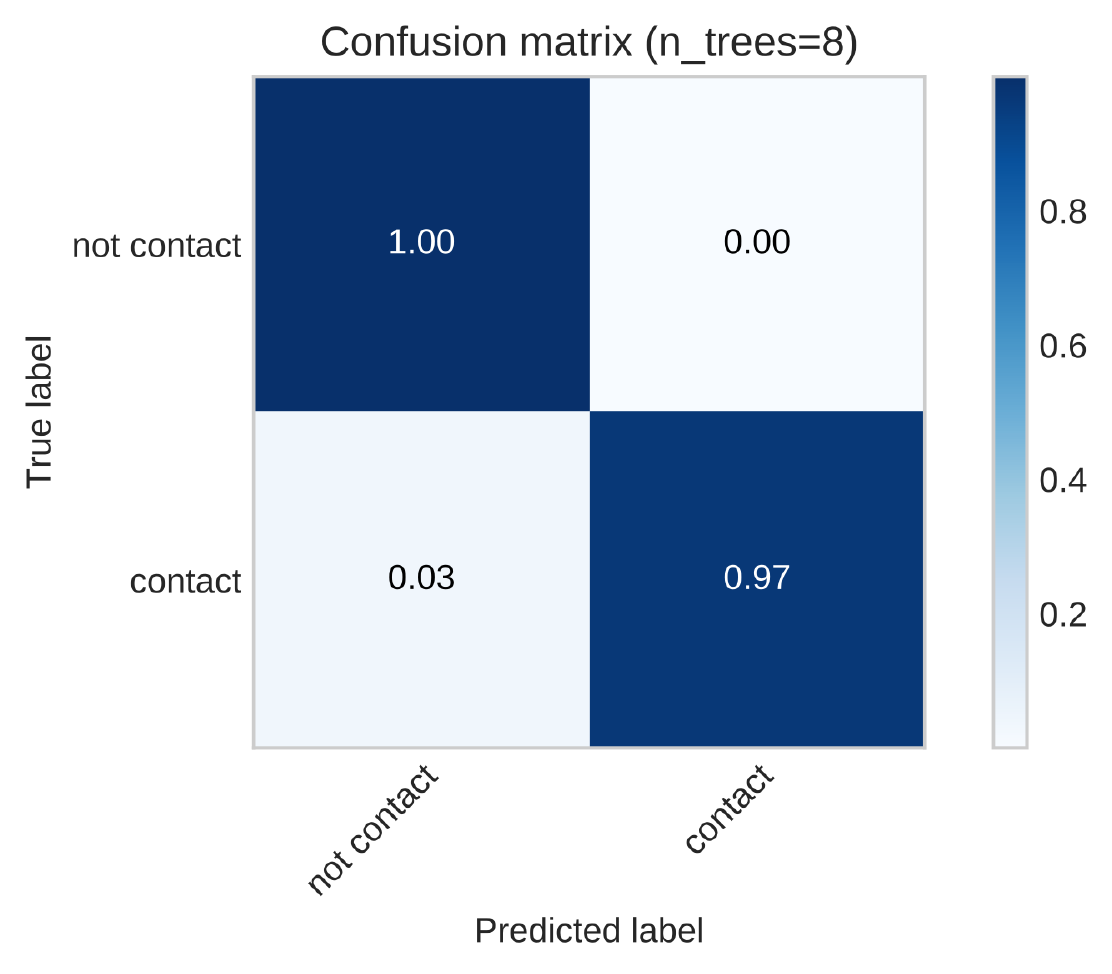}
	}}
	
	\caption{Confusion matrix.}
	\label{fig:confusion_matrix}
\end{figure}

We conducted an experiment to evaluate the extent to which a random forest classifier can detect contact and non-contact events from auditory and force features. The classification results over the test data set are shown in Figure \ref{fig:confusion_matrix} and Table \ref{table:classification_results}.

%%%%%%%%%%%%%%%%%%%%%%%%%%%%%%%%%%%%%%%%%%%%%%%%%%%%%%%%%%%

% % % % % % % % % % % % % %
%CONFUSION MATRIX

\begin{table}[h]
	\caption{Classification results}
	\label{table:classification_results}
	\begin{center}
		\begin{tabular}{c c c c c }
			\cline{2-5}
			 & precision & recall & f1-score & support\\
			\hline
			not contact  &     1.00  &    1.00  &    1.00   &  81368 \\
			contact  & 0.95 & 0.97 & 0.96 & 146 \\
			\hline
			weighted average & 1.00 & 1.00 & 1.00 & 81514\\
			\hline
		\end{tabular}
	\end{center}
\end{table}

%FEATURE IMPORTANCE

\begin{figure}[thpb]
	\centering
	\framebox{\parbox{3in}{
			\includegraphics[scale=0.5]{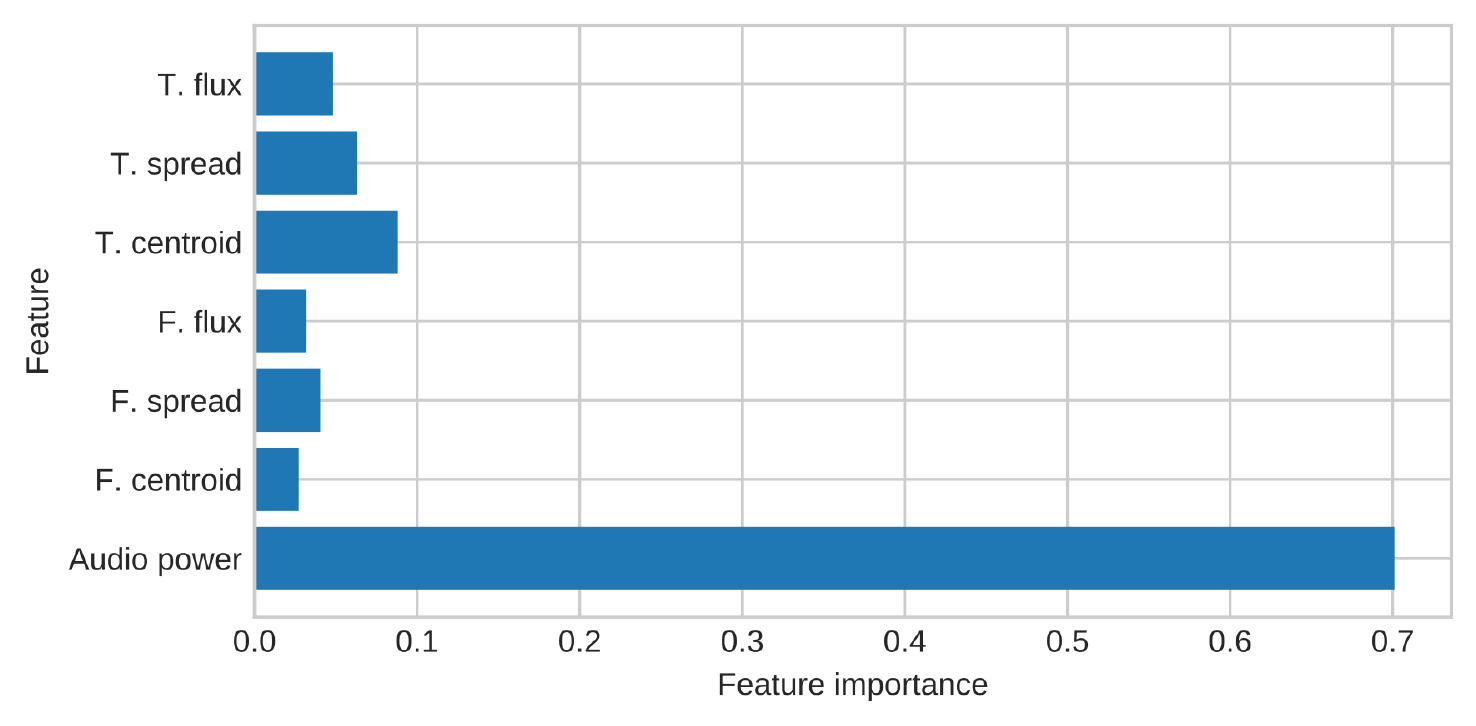}
	}}
	
	\caption{Importance of audio, force (F) and torque (T) features.}
	\label{fig:feature_importance}
\end{figure}

In addition to the confusion matrix and the classification results shown in Table \ref{table:classification_results}, the random forest algorithm provides the \textit{feature importance}, a summary statistic which quantifies how informative each feature is for the classification task. The \textit{feature importance} is expressed as a number between 0 and 1  which describes the importance of each feature for the classification decision. The importance of all features sums to 1. The feature importance results shown in Figure \ref{fig:feature_importance} indicate that the \textit{audio power} is the most informative feature in order to detect contact events, followed by the \textit{torque centroid} and the \textit{torque spread}. 

%%%%%%%%%%%%%%%%%%%%%%%%%

\section{Conclussions and discussion}
We have introduced an approach to detect contacts during manipulation tasks using auditory and force signals. We selected time and frequency domain features that represent the signature characteristics of contacts in the auditory and haptic modalities that have been reported in the literature. We used these features to train a classifier model. The classification results indicate that the features are appropriate for the multimodal detection of contacts. Our results indicate that contact detection is possible independent of the use of force/torque threshold values and we provide an alternative method to the use of filters to process the force/torque signals.

Regarding the auditory modality, our results show that the sound information is important for the detection of contacts (as quantified by the \textit{feature importance} shown in Figure \ref{fig:feature_importance}) and that, combined with force/torque features, is robust against exogenous contact sounds that would otherwise generate false detections. The low feature importance of the force and torque features might indicate that they encode redundant information \cite{Mueller_Guido_2017}. Therefore, classification with a reduced set of these features should be investigated in order to assess whether similar detection results can be achieved with less features. 

We used frequency domain features of the force/torque signals to detect contacts. The use of these features can be extended in order to identify segments of task execution. A visual inspection of Figure \ref{fig:features_over_all_trials} suggests that movement toward and away from the table, as well as the sustained contact of the glass over the table can be identified based on the spectral features. This would enable task monitoring and the identification sub-task completions independent of the force or torque threshold values.

\section*{Acknowledgements}
The research reported in this paper has been supported by the German Research Foundation, as part of Collaborative Research Center 1320 "EASE - Everyday Activity Science and Engineering", University of Bremen (http://www.ease-crc.org/). The research was conducted in subproject H01 "Acquiring activity models by situating people in virtual environments" and subproject R01 "NEEM-based embodied knowledge framework". The authors would like to thank Thorsten Kluss for his support during the data acquisition.

%
%
%
%
% ---- Bibliography ----
%
% BibTeX users should specify bibliography style 'splncs04'.
% References will then be sorted and formatted in the correct style.
%
\bibliographystyle{splncs04}
\bibliography{robot_acoustic_impact}
\end{document}